\documentclass[10pt,twocolumn,letterpaper]{article}
\usepackage[pagenumbers]{cvpr}
\usepackage{graphicx}
\usepackage{amsmath,amssymb}
\usepackage{booktabs,multirow,tabularx}
\usepackage{array}
\usepackage{microtype}
\usepackage{xcolor}
\DeclareRobustCommand{\finalrev}[1]{#1}

\definecolor{cvprblue}{rgb}{0.21,0.49,0.74}
\definecolor{pendingcolor}{rgb}{0.55,0.29,0.08}
\usepackage[pagebackref,breaklinks,colorlinks,allcolors=cvprblue]{hyperref}
\usepackage[capitalize]{cleveref}

\newcommand{\R}{\mathbb{R}}

\newcommand{\Z}{\mathbf{Z}}

\newcommand{\Q}{\mathbf{Q}}

\newcommand{\LN}{\operatorname{LN}}
\newcommand{\CA}{\operatorname{CA}}
\newcommand{\SA}{\operatorname{SA}}
\newcommand{\MSE}{\operatorname{MSE}}

\graphicspath{{figures/}}

\newcommand{\method}{APLaW-V2X}

\makeatletter
\renewcommand\LenToUnit[1]{\strip@pt\dimexpr#1\relax}
\makeatother

\def\confName{CVPR}
\def\confYear{2027}
\title{APLaW-V2X: Asynchronous Persistent Latent Worlds\\for Cooperative Generative Planning}
\author{Zhiguo Zhao \quad Siyi Yan \quad Peitong Wu \quad Xiaoxi Hu \quad Kun Chen\\
Cong Zhao \quad Cairong Zhao \quad Yuchuan Du \quad Yuxiong Ji}
\begin{document}
\maketitle
\suppressfloats[t]
\begin{figure}[t]
\centering
\begin{subfigure}{\columnwidth}
\centering
\input{figures/intro_comparison/panel_a_architectures.tex}
\caption{Cooperation mechanisms.}
\label{fig:intro_architectures}
\end{subfigure}
\par\vspace{4pt}
\begin{subfigure}{\columnwidth}
\centering
\includegraphics[width=\linewidth]{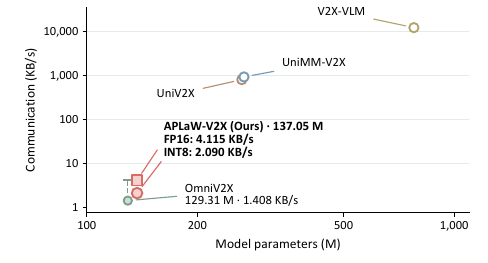}
\caption{Model size and communication cost.}
\label{fig:intro_costs}
\end{subfigure}
\caption{\textbf{Cooperation mechanisms and resource costs.}
Infra. denotes infrastructure-side.
\textit{(a)} Source-Time Replay maintains the current world for trajectory planning; future prediction supplies training supervision (dotted).
\textit{(b)} Log axes; open/filled markers denote reported/counted sizes~\cite{dhvlm}. The retained APLaW parameter count (137.05\,M) includes the auxiliary future predictor.
OmniV2X's segment spans packed 1.408 and JSON 4.231\,KB/s encodings~\cite{omniv2x}. APLaW's INT8/FP16 rates (circle/square) include the 33-byte packet header. Rates follow the respective application-message accounting; 1\,KB=1,000\,B.}
\label{fig:intro_comparison}
\end{figure}

\begin{abstract}
Roadside observations provide complementary information for
autonomous driving, yet communication delays may cause valuable
evidence to arrive after the observed scene has changed.
This raises two key questions: whether delayed roadside
information can still benefit current planning and how to
efficiently communicate planning-relevant information under
limited bandwidth.
To address these challenges, we propose \method, a cooperative
planning framework built upon a persistent shared latent world.
First, we maintain recurrent world states that continuously
accumulate historical observations and introduce Source-Time
Replay to assimilate delayed roadside evidence at its original
capture time, propagating historical corrections to the
current planning state.
Second, we develop Planning-Oriented Communication (POC),
which extracts compact Latent State Tokens (LSTs) from
persistent roadside memory to reduce communication overhead
while preserving information essential for trajectory planning.
The resulting world representation supports trajectory
generation and is further enhanced by action-conditioned
future prediction.
Extensive experiments on DAIR-V2X-Seq demonstrate that
\method\ achieves 90.91 PDMS and 0.83\,m average L2 error
with only 1,057-byte roadside messages.
Comprehensive ablations further validate the effectiveness
of persistent world modeling and planning-oriented
latent state communication.
\end{abstract}

\section{Introduction}
\label{sec:introduction}

Roadside sensing extends autonomous driving beyond the ego vehicle's
field of view, revealing crossing traffic, distant approaches, and
participants hidden by occlusion. Recent end-to-end cooperative driving
systems have increasingly connected such observations to trajectory
planning: UniV2X integrates cooperative perception and planning~\cite{univ2x},
UniMM-V2X develops multi-level cooperative reasoning~\cite{unimmv2x},
and OmniV2X conditions generative planning on vehicle and
infrastructure observations~\cite{omniv2x}. In practice, however,
roadside evidence may arrive after the scene it describes has changed.
This raises our first question: \emph{Can delayed roadside information
still benefit current planning, and how should it be used?}

Consider a vehicle approaching an intersection. A roadside camera may
observe a crossing participant before that participant becomes occluded
from the ego view. Even if this observation reaches the vehicle later,
it can help establish where the participant was and how the traffic
scene has evolved. The value of roadside evidence therefore depends on
both its content and its place in the scene's history. Each packet has
a capture time, identifying the observed event, and an arrival time,
determining when the vehicle can use it. A planning state must retain
past evidence across successive decisions and incorporate a late
observation at the stage of evolution to which it belongs. Work on
collaborative temporal alignment, such as SyncNet~\cite{syncnet},
highlights the role of historical cooperation; latent driving world
models further provide a natural representation for advancing scene
states through time~\cite{law,latentwam,driveworldvla,wote}.

Beyond temporal alignment, communication overhead also affects the
timeliness of roadside information. Transmitting dense spatial features
requires substantial bandwidth and may introduce additional latency,
motivating our second question:
\emph{Can compact latent state tokens efficiently convey planning-relevant
information while reducing communication overhead and latency?}
Existing methods exchange object-level predictions or compressed spatial
features~\cite{where2comm}. For instance, V2X-WAM transmits quantized infrastructure
features and employs learned scene latents after cooperative
fusion~\cite{v2xwam}.

To address these two questions, we present \method, a cooperative
planning framework organized around a \emph{persistent shared latent
world} (Fig.~\ref{fig:intro_comparison}). Vehicle and infrastructure
agents maintain their own recurrent world states in a common latent
schema. A geometry-aware encoder associates visual evidence with
camera rays in a shared reference frame, while learned state
transitions advance each memory using elapsed time and executed motion.

Within this persistent world, \emph{Source-Time Replay} is the key
mechanism for assimilating delayed roadside evidence. Upon receiving a
packet, the vehicle restores the cached state preceding its capture
event, integrates the new evidence at that event, and replays subsequent
state transitions and observations up to the current decision. The
historical correction is thereby propagated through the intervening
scene evolution. A bounded event history confines replay to the
retained interval, while recurrent ego updates continue to advance the
world state between roadside messages.

For efficient cooperation, we introduce \emph{Planning-Oriented
Communication} (POC) through \emph{Latent State Tokens} (LSTs).
Learned queries extract a small set of tokens from the accumulated
roadside memory; the tokens are compressed into a fixed-size INT8
message and decoded into the vehicle's latent world. World-modeling
objectives and planning gradients jointly shape this communication
representation so that its limited capacity carries information
useful for driving decisions. The updated world state provides planning-relevant context for
trajectory generation, while action-conditioned future prediction
further enhances the latent representation.

Our main contributions are summarized as follows:
\begin{itemize}
    \item We propose a \emph{persistent shared latent world} for
    asynchronous cooperative planning, enabling continuous
    accumulation of historical evidence and effective integration
    of delayed roadside observations through Source-Time Replay.

    \item We develop \emph{planning-oriented latent state communication}
    that efficiently exchanges compact Latent State Tokens (LSTs),
    preserving planning-relevant information with substantially
    reduced communication overhead.

    \item Extensive experiments on DAIR-V2X-Seq demonstrate the
    effectiveness of \method, achieving 90.91 PDMS with only
    1,057-byte roadside messages. Comprehensive ablations further
    validate the benefits of persistent world modeling,
    Source-Time Replay, and latent state communication.
\end{itemize}

\section{Related Work}
\label{sec:related_work}

\paragraph{V2X cooperative planning.}
UniV2X integrates cooperative perception and planning through
sparse--dense vehicle--infrastructure communication~\cite{univ2x}.
UniMM-V2X extends cooperation across perception and prediction using
multi-level, mixture-of-experts-enhanced fusion~\cite{unimmv2x}.
OmniV2X conditions generative planning on independently encoded
multimodal and infrastructure contexts~\cite{omniv2x}.
V2X-WAM combines quantized roadside feature communication with
action-conditioned future occupancy and dynamic-flow prediction for
trajectory refinement~\cite{v2xwam}.
For asynchronous collaboration, SyncNet leverages historical cooperative
information to compensate latency in collaborative perception~\cite{syncnet}.
\method\ maintains persistent cooperative world states and assimilates
delayed evidence at its capture time through Source-Time Replay.

\paragraph{Latent world models for driving.}
LAW predicts future latent scene features conditioned on current features
and ego trajectories to improve driving representations~\cite{law}.
Latent-WAM compresses multi-view images into compact scene tokens and
models their dynamics from visual and motion histories~\cite{latentwam}.
DriveWorld-VLA integrates latent-space world modeling with action
planning through shared world-model states~\cite{driveworldvla}.
\method\ extends latent world modeling to asynchronous V2X cooperation
through recurrent states and planning-oriented tokens.

\begin{figure*}[t]
    \centering
    \includegraphics[width=\textwidth]{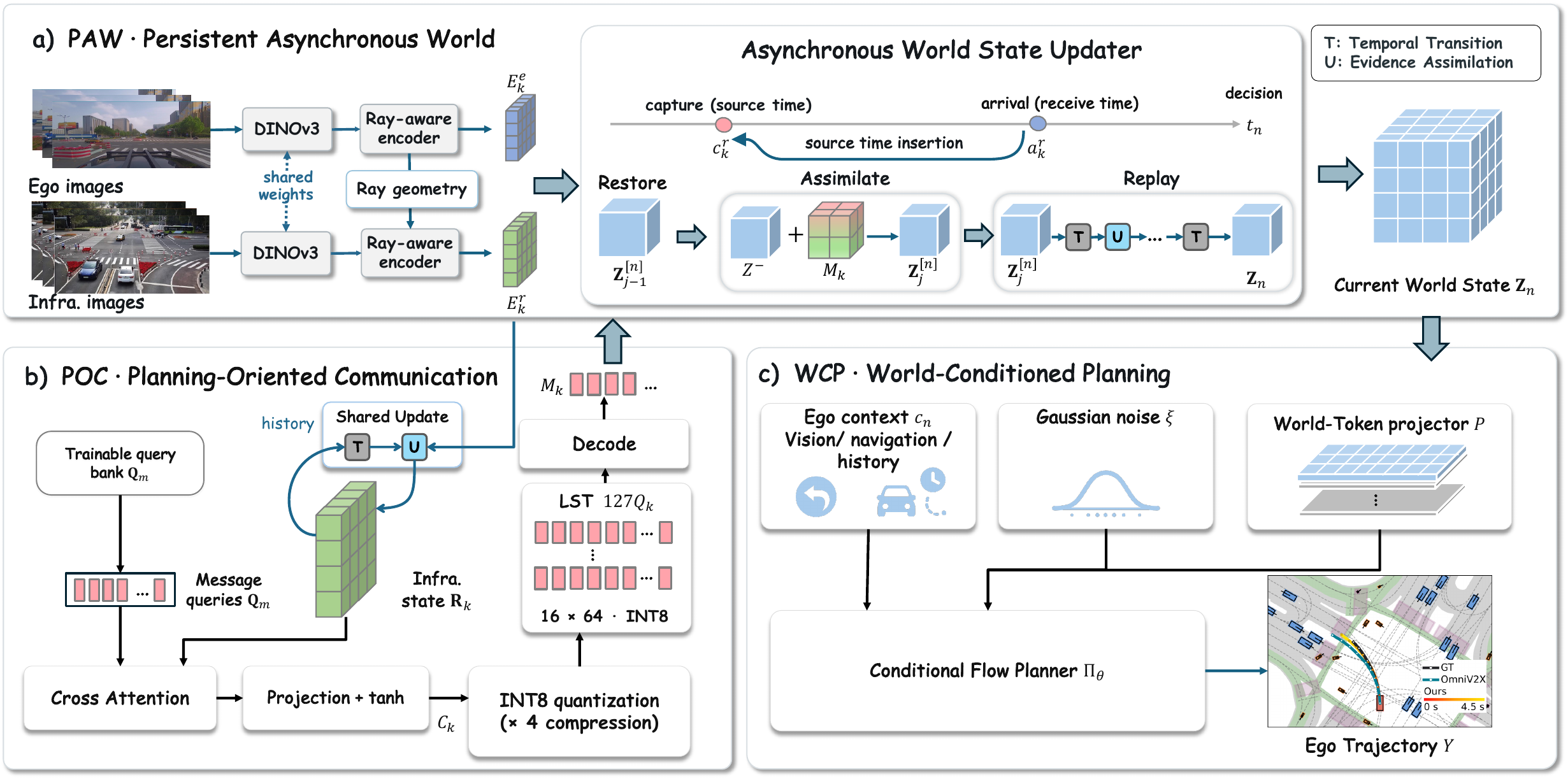}
    \caption{\textbf{\method\ overview.}
    Ray-aware encoders produce observation slots $E_k^e,E_k^r$.
    Learned message queries $\Q_m$ aggregate $\mathbf R_k$ into codes
    $C_k$, transmitted as INT8 $127Q_k$ and decoded to $M_k$.
    After arrival $a_k^r\le t_n$, delayed tokens are inserted at the capture
    event $s_j=c_k^r$; replay reconstructs $\Z_j^{[n]}$ through the current
    state $\Z_n$. Projected world tokens condition native trajectory
    generation; action-conditioned prediction supplies training-only
    future-state supervision.}
    \label{fig:framework}
\end{figure*}

\section{Method}
\label{sec:method}

\subsection{Framework Overview}
\label{sec:formulation}
\method\ comprises three modules (\finalrev{Fig.}~\ref{fig:framework}):
\emph{Persistent Asynchronous World} (PAW) maintains geometrically
conditioned memories and incorporates delayed evidence through
Source-Time Replay; \emph{Planning-Oriented Communication} (POC)
compresses roadside memory into task-supervised messages; and
\emph{World-Conditioned Planning} (WCP) generates trajectories from
the current vehicle state and native ego context.

At decision time $t_n$, the vehicle has observed images, calibration,
executed odometry, and native context $c_n$ containing ego visual,
navigation, and trajectory-history information. A roadside packet has
capture time $c_k^r$ and arrival time $a_k^r$. Only packets with
$a_k^r\leq t_n$ are admitted; a lost packet has $a_k^r=\infty$.
Arrival time determines availability, whereas capture time locates
evidence in the state history. Capture times remain on their original
sampling grids.

Continuous evaluation fixes a reference $\mathcal R$ at each stream's
first ego pose and carries both memories forward. Training uses history
windows anchored at the current ego pose. Camera rays and executed
motion are expressed in $\mathcal R$; the output
$Y\in\R^{9\times3}$ contains positions and headings in the current
ego frame at nominal $0.5$\,s intervals.

\subsection{Persistent Asynchronous World (PAW)}
\label{sec:paw}
\label{sec:representation}
\label{sec:async}
\paragraph{Geometry-aware shared representation.}
The vehicle state $\Z_n$ and roadside state $\mathbf R_k$ each contain
$K=64$ latent slots of width $d=256$. They share a slot parameterization,
geometric reference, and learned update operators, while retaining
their respective observation histories.
\finalrev{Bold symbols denote world states and learned queries;
other tensors are italic, with tokens stored row-wise.}
A frozen DINOv3 backbone~\cite{dinov3}, reused from OmniV2X, produces
784 spatial features of width 768 per $448\times448$ image. For source
$v\in\{e,r\}$, patch $i$ has feature $f^v_{ki}$, ray origin $o^v_{ki}$,
and unit direction $r^v_{ki}$ in $\mathcal R$. The ray-aware encoder
combines appearance, geometry, and source identity:
\begin{equation}
 h^v_{ki}=W_f\LN(f^v_{ki})+
 \phi_g([o^v_{ki}/s_g,r^v_{ki}])+e_v,
 \label{eq:observation}
\end{equation}
\finalrev{Here $W_f\in\R^{256\times768}$ projects appearance, $s_g=50$\,m,
$\phi_g$ is a two-layer multilayer perceptron (MLP), and $e_v$ is a source embedding.}

Two cross-attention blocks read these features using shared learned
queries $\Q\in\R^{K\times d}$, producing observation slots $E_k^v$.
CA and SA denote residual cross- and self-attention blocks with
feed-forward layers; LN denotes layer normalization. Invalid patches
are masked, and an always-valid null token handles empty observations.
The world branch reads spatial patch tokens; the native visual branch
retains its original token layout. Both states start from $\Q$ and
accumulate available evidence through shared operators. Rays and attention
weights provide spatial information, while two-view reconstruction and
full-information teacher matching train the shared slots to retain
complementary evidence.

\paragraph{Temporal transition and evidence assimilation.}
PAW separates temporal transition $T$, which advances memory to the
next event time, from evidence assimilation $U$, which incorporates
available observations. Both chronological updates and replay use these
operators. Let $u_k=(\Delta x_{\mathcal R},\Delta y_{\mathcal R},\Delta\psi)$
be executed motion between updates separated by $\delta_k$:
\begin{equation}
 T(\Z,u_k,\delta_k)=\Z+
 \finalrev{\delta_k\SA\!\left(\Z+\phi_u(\bar u_k,\delta_k)\right)W_T},
 \label{eq:transition}
\end{equation}
\finalrev{Here $\bar u_k$ rescales XY by $s_g$ and retains yaw;
$\phi_u$ embeds motion and elapsed time, broadcast over slots,
and $W_T\in\R^{256\times256}$ projects features.} Elapsed time scales the
residual, making $T$ the identity at $\delta_k=0$. The stationary
roadside agent uses zero platform motion and its recorded elapsed time.

The propagated prior $\Z^-$ first reads local observation slots and
then decoded roadside tokens $M_k$ from POC:
\begin{gather}
 \widetilde{\Z}=\operatorname{select}_{b_e}(\CA_e(\Z^-,E_k^e),\Z^-),\nonumber\\
 \Z^+=\operatorname{select}_{b_r}(\CA_r(\widetilde{\Z},M_k),\widetilde{\Z}),\nonumber\\
 U(\Z^-,E_k^e,M_k)=\LN(\SA(\Z^+)).
 \label{eq:update}
\end{gather}
Here $\operatorname{select}_b(x,y)$ returns $x$ when evidence is present
and $y$ otherwise. Self-attention mixes slots before normalization.
An ego-only event sets $b_r=0$; a remote-only event sets $b_e=0$.
The roadside state uses the same transition and assimilates its own
observation through the remote-attention block before supplying its
updated memory to POC.

\paragraph{Source-Time Replay.}
At decision $t_n$, the receiver orders observed ego captures and
admitted roadside captures into source-time events $s_j$. Let
$\Z_j^{[n]}$ denote the state at event $j$, reconstructed from
information available by $t_n$:
\begin{equation}
 \Z_j^{[n]}=U\!\left(T(\Z_{j-1}^{[n]},u_j,s_j-s_{j-1}),E_j^e,M_j\right).
 \label{eq:replay}
\end{equation}
Here $E_j^e$ and $M_j$ reindex $E_k^e$ and $M_k$ by
source event $s_j$; missing inputs have zero validity flags.
The final event $J_n$ is the current ego tick, $s_{J_n}=t_n$, giving
the planning state $\Z_n=\Z_{J_n}^{[n]}$.
Upon a delayed arrival, PAW inserts decoded tokens at $c_k^r$ and
restores the preceding cached state. It advances to capture time,
assimilates the message, and replays subsequent motions and observations
through $t_n$. Thus, delayed evidence corrects its source-time state
before the correction reaches the current decision. Batched late
messages trigger recomputation from the earliest changed event;
co-timed messages follow sequence-number order.

The buffer retains eight ego poses, intermediate roadside events, a
predecessor state, and cached observation encodings. For fixed weights,
predecessor, and ordered admitted events, replay reproduces the
chronological recurrence, with work proportional to the affected retained
events. Between arrivals, temporal transition and ego assimilation
continue updating the planning state. Appendix~B details interpolation,
packet admission, the event algorithm, and the recurrence argument.

\subsection{Planning-Oriented Communication (POC)}
\label{sec:communication}
\paragraph{Query-based compression and quantization.}
POC converts accumulated roadside memory into compact messages for PAW.
Learned queries $\Q_m\in\R^{16\times256}$ aggregate $\mathbf R_k$;
a channel projection and bounded activation produce codes that are
quantized and decoded as
\begin{gather}
 \finalrev{C_k=\tanh(\CA_m(\Q_m,\mathbf R_k)W_c)},\nonumber\\
 Q_k=\operatorname{round}(127C_k)/127,\quad \finalrev{M_k=Q_kW_d+P_m}.
 \label{eq:codec}
\end{gather}
\finalrev{Learned projections $W_c\in\R^{256\times64}$ and
$W_d\in\R^{64\times256}$ yield $16\times64$ codes $C_k,Q_k$ and
$16\times256$ decoded tokens $M_k$; $P_m$ contains learned
message-position embeddings.} Each packet summarizes roadside
evidence accumulated through its capture time.

Training uses a straight-through estimator; deployment serializes
$127Q_k$ as INT8, with 1,024 payload bytes and a 33-byte header.
Packets are independently decoded. This codec supplies all roadside
evidence in the latent-only configuration; Appendix~J gives
implementation details.

\paragraph{Planning-oriented message learning.}
World and planning losses supervise messages through decoding and
receiver updates. Learned aggregation summarizes recurrent roadside
memory, while planning gradients reward trajectory-relevant summaries
and teacher matching supervises the vehicle representation.
Section~\ref{sec:ablation_studies} evaluates recurrent-memory
aggregation and planning supervision at the same packet size.

\subsection{World-Conditioned Planning (WCP)}
\label{sec:planning}
\label{sec:future}
\paragraph{Current-world conditioned flow planning.}
WCP augments OmniV2X's native context with the current PAW state.
The world-token projector $P$ applies a current-world type embedding,
layer normalization, a $256\to384$ linear map, and output normalization.
The resulting 64 tokens are appended after native context encoding at
each of the 16 UnifiedDiT blocks, retaining their learned slot
identities in joint cross-attention. \finalrev{The native planner $\Pi_\theta$,
with parameters $\theta$ and Gaussian noise $\xi$, generates}
\begin{equation}
 Y=\Pi_\theta([c_n,P(\Z_n)];\xi).
 \label{eq:planning}
\end{equation}
Each decision uses one Flow Planner invocation with 20 flow integration
steps. Ego visual features are extracted once, and the reconstructed
current state supplies world context throughout sampling. The native
flow learns coordinate-space velocities between Gaussian noise and
expert labels $D^\star$ (XY displacements and ego-relative
headings)~\cite{flowmatching,rectifiedflow}. Output conversion to $Y$ accumulates XY
displacements and normalizes headings; Appendix~J specifies training
interpolation and label units.

\paragraph{Training-only future-world supervision.}
An action-conditioned predictor encourages predictive structure in the
world representation, while a future-conditioned planning objective
connects it to trajectory quality. Given candidate positions $p_h$,
actions are $A_h=R_{\psi_n}(p_h-p_{h-1})/s_a$, with $s_a=10$\,m and
$p_0$ the native current position. \finalrev{$R_{\psi_n}$ rotates displacements
by the current yaw $\psi_n$ into $\mathcal R$.} Action embeddings, waypoint-position embeddings, and
current-pose embedding $\rho_n$ form tokens $B$, from which the predictor
produces a terminal state at $\tau=4.5$\,s\finalrev{, with parameters $\mu$}:
\begin{equation}
 \widehat\Z_f=F_\mu(\Z_n,A,\rho_n)
 =\finalrev{\Z_n+\SA_\mu(\CA_\mu(\Z_n,B))W_\mu}.
 \label{eq:future}
\end{equation}
\finalrev{$W_\mu\in\R^{256\times256}$ projects the attention output.}
Expert actions $A^\star$ condition future-state supervision
through $\widehat\Z_f^\star$; sampled proposals condition
the auxiliary planning term below through
$\widehat\Z_f^{\rm prop}$. This branch is
used only during training: inference generates trajectories directly
from the current world state as in Eq.~\eqref{eq:planning}.

\subsection{Training Objectives and Optimization}
\label{sec:objectives}
\paragraph{World representation objectives.}
A teacher encodes synchronized, uncompressed roadside information to
obtain current target $\bar\Z_n$. It propagates its vehicle and roadside
states and assimilates actual future observations to obtain
$\bar\Z_f$; both targets are detached. A geometry-query decoder
reconstructs normalized frozen backbone features from the current and
expert-action-conditioned future states. Losses $\mathcal L_{\rm rec}$
and $\mathcal L_{\rm frec}$ average squared error over valid patches
and channels and sum the two views:
\begin{align}
 \mathcal L_{\rm world}={}&\mathcal L_{\rm rec}
 +\MSE(\widehat\Z_f^\star,\bar\Z_f)
 +\mathcal L_{\rm frec}\nonumber\\
 &+\tfrac12\MSE(\Z_n,\bar\Z_n)
 +\tfrac14\mathcal L_{\rm trans}.\label{eq:worldloss}
\end{align}
\finalrev{MSE denotes mean squared error.} Transition loss compares the prior propagated from the previous history
endpoint with $\bar\Z_n$ and is zero for one-frame histories.
Reconstruction ties slots to visual evidence, teacher matching aligns
compressed and full-information states, and transition supervision
trains state evolution between observations.

\paragraph{Planning objectives.}
\finalrev{Current-world and auxiliary future-world conditioning use
the native flow-matching loss $\mathcal L_{\rm native}$:}
\begin{align}
 \mathcal L_{\rm plan}={}&\tfrac12\mathcal L_{\rm native}(D^\star\mid c_n,\Z_n)\nonumber\\
 &+\tfrac12\mathcal L_{\rm native}(D^\star\mid c_n,\Z_n,\widehat\Z_f^{\rm prop}).
 \label{eq:planloss}
\end{align}
For the second term, a stopped-gradient native sample from the current
state provides a proposal whose actions condition $F_\mu$.
The future-conditioned loss backpropagates through predictor and planner
conditioning paths to the current state, codec, and projector.
Proposal sampling is detached; conditioning paths remain differentiable
even when predictor and native planner parameters are frozen.

\paragraph{Staged optimization.}
Training first learns world modules from synchronous full roadside
states with an \finalrev{exponential moving average (EMA)} teacher (decay 0.99). Stage two optimizes
the quantized asynchronous path using
$\mathcal L_{\rm world}+\mathcal L_{\rm plan}$,
with the teacher and native planner fixed.
This stage trains the codec, state updates, and context projector.
Stage three jointly fine-tunes these modules to align world
representations, asynchronous updates, communication, and planning.
The visual backbone remains frozen.
Evaluation uses the final stage-three checkpoint.
\finalrev{See the appendix.}

\section{Experiments}
\label{sec:experiments}

\begin{table*}[!t]
\begin{minipage}{\textwidth}
\centering
\caption{Planning quality and communication cost on DAIR-V2X-Seq.
OmniV2X ($\ddagger$) is reproduced from released checkpoints.
APLaW-V2X and reproduced OmniV2X average 10 inference seeds
for L2/collision and five for PDMS.
Other methods use published results~\cite{unimmv2x,v2xwam}.
UniMM-V2X retains its benchmark horizons; V2X-WAM ($\P$)
uses a map and waypoint-averaged L2.
OmniV2X: original object-message and map-message accounting;
$\dagger$: measured application-layer bytes/s with headers.
Bold marks our results and tied collision rates.}
\label{tab:main}
\begingroup
\footnotesize
\setlength{\tabcolsep}{2.6pt}
\renewcommand{\arraystretch}{1.16}
\begin{tabular*}{\textwidth}{@{\hspace{14pt}\extracolsep{\fill}}lccrrrrrr@{}}
\toprule
\multirow{2}{*}{\textbf{Method}} & \multirow{2}{*}{\shortstack{\textbf{V2X Message}\\\textbf{Type}}} & \multirow{2}{*}{\shortstack{\textbf{Trans. Cost}\\\textbf{(B/s)} $\downarrow$}} & \multicolumn{4}{c}{\textbf{L2 Error (m)} $\downarrow$} & \multirow{2}{*}{\shortstack{\textbf{3s Avg. Collision}\\\textbf{Rate (\%)} $\downarrow$}} & \multirow{2}{*}{\textbf{PDMS} $\uparrow$}\\
\cmidrule(lr){4-7}
& & & 1\,s & 2\,s & 3\,s & Avg. & & \\
\midrule
\llap{\multirow{3}{*}{\rotatebox[origin=c]{90}{\scriptsize\textit{Ego-Only}}}\hspace{6pt}}VAD~\cite{vad} & -- & -- & 1.65 & 2.72 & 3.80 & 2.72 & 1.12 & --\\
UniAD~\cite{uniad} & -- & -- & 1.26 & 2.22 & 3.06 & 2.18 & 1.13 & --\\
SparseDrive~\cite{sparsedrive} & -- & -- & 1.02 & 1.69 & 2.37 & 1.69 & 0.99 & --\\
\midrule
\llap{\multirow{7}{*}{\rotatebox[origin=c]{90}{\scriptsize\textit{Coop V2X}}}\hspace{6pt}}V2VNet~\cite{v2vnet} & BEV Features & $8.19\times10^{7}$ & 1.96 & 2.37 & 3.41 & 2.58 & 0.88 & --\\
Coopernaut~\cite{coopernaut} & BEV Features & $8.19\times10^{7}$ & 2.69 & 4.07 & 5.50 & 4.09 & 1.42 & --\\
UniV2X~\cite{univ2x} & Queries \& Occ. & $8.09\times10^{5}$ & 1.45 & 2.19 & 3.04 & 2.23 & 0.25 & --\\
UniMM-V2X~\cite{unimmv2x} & Queries \& Occ. & $9.32\times10^{5}$ & 0.78 & 1.63 & 2.05 & 1.49 & 0.12 & --\\
V2X-WAM$^{\P}$~\cite{v2xwam} & INT8 BEV features & 98,400 & 0.53 & 0.92 & 1.45 & 0.97 & \textbf{0.01} & --\\
\addlinespace[2pt]
OmniV2X (no map)$^{\ddagger}$ ~\cite{omniv2x} & Object tokens & 1,408 & 0.48 & 0.86 & 1.32 & 0.89 & 0.09 & 88.70\\
OmniV2X (w/ map)$^{\ddagger}$ ~\cite{omniv2x} & Object \& map tokens & 25,792 & 0.48 & 0.88 & 1.34 & 0.90 & \textbf{0.01} & 90.01\\
\midrule
\addlinespace[2pt]
\llap{\multirow{1.4}{*}[2pt]{\rotatebox[origin=c]{90}{\scriptsize\textit{Ours}}}\hspace{6pt}}APLaW-V2X & Latent State Tokens & \textbf{2,090.1}$^{\dagger}$ & \textbf{0.43} & \textbf{0.82} & \textbf{1.23} & \textbf{0.83} & \textbf{0.01} & \textbf{90.91}\\
\textcolor{gray}{\textbf{APLaW-V2X (FP16)}} & \textcolor{gray}{\textbf{FP16 Latent Tokens}} & \textcolor{gray}{\textbf{4,114.9}$^{\dagger}$} & \textcolor{gray}{\textbf{0.42}} & \textcolor{gray}{\textbf{0.81}} & \textcolor{gray}{\textbf{1.22}} & \textcolor{gray}{\textbf{0.82}} & \textcolor{gray}{\textbf{0.01}} & \textcolor{gray}{\textbf{91.02}}\\
\addlinespace[2pt]
\bottomrule
\end{tabular*}
\endgroup
\end{minipage}
\end{table*}

\subsection{Experimental Setup}
\label{sec:experimental_setup}

\paragraph{Dataset and evaluation.}
We use DAIR-V2X-Seq~\cite{v2xseq}, the UniV2X scene split, and the planning evaluation interface adopted by OmniV2X~\cite{omniv2x}.
The split contains 5,391 training windows from 43 recording sources and 2,390 validation planning anchors from 19 disjoint sources.
Validation anchors form 95 chronological streams, each starting with fresh vehicle and infrastructure states that accumulate observations over time.
Our ablations and robustness evaluations share observations, recorded poses, communication traces, planning horizons, and evaluation metrics.

\paragraph{Metrics and communication accounting.}
We report Predictive Driver Model Score (PDMS; higher is better), L2 trajectory error and average collision rate (lower is better), and application-layer communication cost under the adapted OmniV2X evaluator~\cite{omniv2x}.
For our evaluations and the reproduced OmniV2X rows, average L2 aggregates errors at the nominal 1, 2, and 3\,s horizons.
PDMS gains are absolute score-point differences.
Message size includes serialized payload and application headers; the measured rate divides transmitted application bytes by stream duration.
Published methods retain their reported communication accounting and horizon conventions (Table~\ref{tab:main}).

\paragraph{Model and inference configuration.}
The cooperative state contains $64\times256$ latent slots.
Each default LST message carries $16\times64$ INT8 values: 1,024 payload bytes and a 33-byte header, totaling 1,057 bytes.
The current-state planning path in Section~\ref{sec:planning} assimilates available ego observations and roadside messages, then conditions the native Flow Planner with 20 integration steps.
The default configuration appears as A6, A10, and C0 in the corresponding ablation groups.

\paragraph{Controlled comparisons.}
\emph{Matched training} independently optimizes structural alternatives under a common procedure.
\emph{Fixed-weight intervention} reuses A6 parameters and changes the indicated evidence source or assimilation operation.
The history table reports both; the codec table identifies
precision switching and matched FP16 training.
All ablations (A1--A10 and C0--C4) report PDMS averaged
over five shared inference seeds, ensuring paired comparisons.

\subsection{Comparison with Existing Planning Methods}
\label{sec:sota_comparison}

\begin{figure*}[!t]
\centering
\input{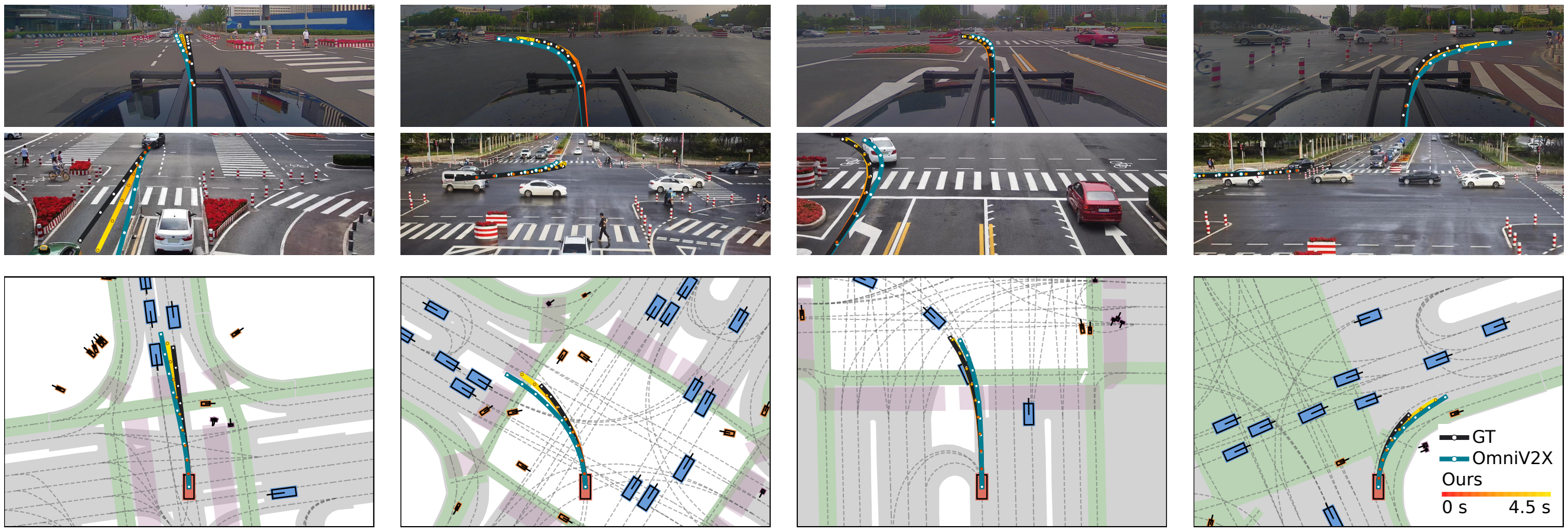}
\caption{Qualitative planning comparisons on four validation scenes.
Each column shows paired vehicle and roadside images above the corresponding bird's-eye view. Ground truth is black, reproduced OmniV2X is cyan, and APLaW-V2X is colored by future time over 4.5\,s. Calibration projects the same trajectories into both camera views. Maps and surrounding agents provide annotation context. Values beneath each scene give PDMS ($\uparrow$) and average L2 in meters ($\downarrow$), in APLaW-V2X / OmniV2X order.}
\label{fig:qualitative}
\end{figure*}

\paragraph{Trajectory quality.}
Table~\ref{tab:main} compares APLaW-V2X with existing
planning methods, including OmniV2X checkpoints reproduced
using ground-truth infrastructure boxes and 16 object tokens.
APLaW-V2X achieves 90.91 PDMS and 0.83\,m average L2,
outperforming map-free OmniV2X by 2.21 PDMS points and
0.06\,m L2, while reducing collision rate from 0.09\%
to 0.01\%. Against map-conditioned OmniV2X, the gains
are 0.90 PDMS points and 0.07\,m L2, with improvements
across all three trajectory horizons. These results demonstrate the effectiveness of APLaW-V2X
in improving trajectory accuracy and reducing evaluated collision rates.

\paragraph{Communication cost.}
APLaW-V2X transmits 2,090.1 application bytes/s with its 1,057-byte packets.
The rate is approximately 97.9\% lower than V2X-WAM's reported 98,400\,B/s and 91.9\% lower than the map-conditioned OmniV2X accounting of 25,792\,B/s.
Map-free OmniV2X uses a compact 1,408\,B/s object-message accounting.
APLaW-V2X achieves state-of-the-art planning performance
with low communication overhead.

\paragraph{Qualitative comparison.}
Figure~\ref{fig:qualitative} presents four selected scenes containing mixed traffic, left turns, and a right turn.
APLaW-V2X achieves higher PDMS and lower L2 than OmniV2X in all four selected cases.
In Fig.~\ref{fig:qualitative}\subref{fig:qualitative_c}, average L2 changes from 0.293 to 0.068\,m and PDMS from 66.7 to 99.9.

\subsection{Ablation Studies}
\label{sec:ablation_studies}

\paragraph{Historical cooperation and temporal assimilation.}
Table~\ref{tab:history_ablation} separates evidence availability (A1/A2), recurrent history (A2/A6), and assimilation timing (A4/A5/A6).
A1 retains the ego state-update path with roadside inputs removed.
A3 substitutes mismatched historical roadside content while retaining the current observation, testing the information consumed by the trained history branch.

\begin{table}[!t]
\centering
\caption{Ablation of Historical Cooperative Memory and Temporal Assimilation. Matched training optimizes each structural alternative; fixed-weight interventions reuse A6 parameters. A3 changes historical content at inference. Higher PDMS is better.}
\label{tab:history_ablation}
\setlength{\tabcolsep}{5pt}
\renewcommand{\arraystretch}{1.12}
\resizebox{\columnwidth}{!}{\begin{tabular}{lcc}
\toprule
\textbf{Configuration} & \textbf{Matched PDMS $\uparrow$} & \textbf{Fixed A6 PDMS $\uparrow$}\\
\midrule
A1: No roadside evidence & 89.39 & 89.43\\
A2: Frame-wise cooperation & 90.00 & 90.60\\
A3: Mismatched historical content & -- & 89.69\\
A4: Arrival-time fusion & 90.41 & 90.30\\
A5: Age-aware fusion & 90.62 & 90.55\\
\midrule
\textbf{A6: History + Source-Time Replay} & \textbf{90.91} & \textbf{90.91}\\
\bottomrule
\end{tabular}}
\end{table}

Matched training gives 89.39 PDMS for A1 and 90.00 for A2, a 0.61-point gain from frame-wise roadside evidence.
Combining persistent history with Source-Time Replay improves PDMS from 90.00 (A2) to 90.91 (A6), a gain of 0.91 points.
Under matched training, Source-Time Replay further outperforms arrival-time and age-aware fusion by 0.50 and 0.29 PDMS points, respectively.

With fixed weights, A6 gains 0.61 points over A4 and 0.36 over A5.
Frame-wise cooperation reaches 90.60, and replacing historical content lowers the score to 89.69, compared with 90.91 for the original history.
This 1.22-point change shows that the trained planner responds to the retained roadside information.

\paragraph{Latent world representation and future supervision.}
Table~\ref{tab:latent_future_ablation} examines the latent state and its training signal.
A7 uses native planner context; A8 adds current-world conditioning; A9 adds action-independent future supervision; A10 conditions that prediction on actions.

\begin{table}[!t]
\centering
\caption{Ablation of Latent World Conditioning and Future Supervision. A7--A10 vary latent conditioning and the future-state training objective.}
\label{tab:latent_future_ablation}
\setlength{\tabcolsep}{4pt}
\renewcommand{\arraystretch}{1.12}
\resizebox{\columnwidth}{!}{\begin{tabular}{llcc}
\toprule
\textbf{ID} & \textbf{Training configuration} & \textbf{PDMS $\uparrow$} & \textbf{\finalrev{Avg.} L2 (m) $\downarrow$}\\
\midrule
A7 & Native context & 89.20 & 0.91\\
A8 & Current-world conditioning & 90.48 & 0.87\\
A9 & Action-independent supervision & 90.74 & 0.85\\
\midrule
\textbf{A10} & \textbf{Action-conditioned supervision} & \textbf{90.91} & \textbf{0.83}\\
\bottomrule
\end{tabular}}
\end{table}

Current-world conditioning raises PDMS from 89.20 to 90.48 and reduces L2 from 0.91 to 0.87\,m.
Action-independent future supervision further raises PDMS to 90.74,
while action-conditioned supervision achieves 90.91 PDMS
and 0.83\,m L2. These results highlight the benefits of latent world modeling
and action-conditioned future supervision for planning.

\paragraph{Planning-oriented communication.}
Table~\ref{tab:codec_comparison} tests aggregation, planning supervision, and precision.
C0/C1/C2 share a 1,057-byte budget: C1 replaces learned aggregation with pooling, and C2 removes the planning-gradient path.
C3a sends FP16 codes using C0 weights; C3b uses matched FP16 training with C0's initialization, architecture, schedule, and supervision.
C4 sends the complete FP16 state.
All variants share the sending schedule and 33-byte header.

\begin{table}[!t]
\centering
\caption{Communication representation and planning quality. Packet sizes include application-layer headers, and L2 is measured in meters. C0--C2 use the same communication budget.}
\label{tab:codec_comparison}
\setlength{\tabcolsep}{3.5pt}
\renewcommand{\arraystretch}{1.12}
\resizebox{\columnwidth}{!}{\begin{tabular}{llrrrr}
\toprule
ID & Representation & B/packet $\downarrow$ & B/s $\downarrow$ & PDMS $\uparrow$ & \finalrev{Avg.} L2 $\downarrow$ \\
\midrule
C0  & Learned + planning, INT8  & 1057  & 2090.1  & 90.91 & 0.83 \\
C1  & Simple pooling, INT8     & 1057  & 2090.1  & 87.31 & 0.97 \\
C2  & Learned, world supervision & 1057 & 2090.1 & 88.56 & 0.95 \\
C3a & FP16, C0 weights         & 2081  & 4114.9  & 89.93 & 0.85 \\
C3b & FP16, matched training   & 2081  & 4114.9  & 91.02 & 0.82 \\
C4  & Full latent state, FP16  & 32801 & 64860.0 & 91.07 & 0.82 \\
\bottomrule
\end{tabular}}
\end{table}

At the common packet budget, learned planning-oriented messages reach 90.91 PDMS, exceeding simple pooling by 3.60 points and representation-based codec training by 2.35 points.
Average L2 falls from 0.97 and 0.95\,m to 0.83\,m, respectively.

Switching the INT8-trained C0 checkpoint directly to FP16 transmission yields 89.93 PDMS (C3a), reflecting sensitivity to the encoding precision.
With matched FP16 training, C3b achieves 91.02 PDMS and 0.82\,m L2.
Compared with C3b, INT8 reduces packet size by 49.21\% while retaining a comparable planning score (90.91 PDMS).
These results demonstrate that planning-oriented LSTs preserve
critical information for trajectory planning while enabling
highly efficient communication.

\subsection{Robustness to Asynchronous Cooperation}
\label{sec:asynchronous_robustness}

\begin{figure}[!t]
\centering
\captionsetup[subfigure]{font=small,skip=2pt,justification=centering,singlelinecheck=false}
\begin{subfigure}[t]{.49\columnwidth}
\centering
\includegraphics[width=\linewidth]{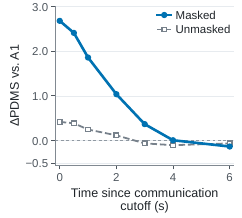}
\caption{Observation conditions}
\label{fig:history_conditions}
\end{subfigure}\hfill
\begin{subfigure}[t]{.49\columnwidth}
\centering
\includegraphics[width=\linewidth]{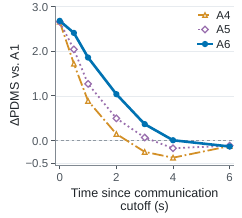}
\caption{Persistence after cutoff}
\label{fig:history_updates}
\end{subfigure}
\begingroup
\begin{minipage}{\columnwidth}
\caption{Persistence of cooperative utility during an outage.
(a) A6--A1 PDMS gain with intact ego inputs and a normalized central image mask $[0.25,0.25,0.75,0.75]$; roadside inputs remain intact. (b) A4/A5/A6 assimilate identical initially delayed evidence from common pre-fusion states. After assimilation at $t_0$, each state evolves with the same ego observations and transitions. Curves report PDMS gains over the matched ego reference against elapsed outage time $\tau=t-t_0$.}
\label{fig:history_age}
\end{minipage}
\endgroup
\end{figure}

\paragraph{Persistence after communication cutoff.}
Figure~\ref{fig:history_age}(a) measures the value of previously received evidence during an outage under intact and centrally masked ego images.
Roadside images remain intact.
Retained cooperative utility is
\begin{equation}
 \Delta_{\rm hist}(\tau)=\operatorname{PDMS}_{\rm coop}(\tau)
 -\operatorname{PDMS}_{\rm ego}(\tau),
 \label{eq:history_utility}
\end{equation}
using the same evaluation cohort and time for the cooperative and ego references.
Positive values identify planning benefits retained from roadside evidence during the interruption.

In panel (b), A4/A5/A6 assimilate the same delayed evidence from identical pre-fusion states and histories under the central ego-image mask.
The outage starts after assimilation at $t_0$, with elapsed time $\tau=t-t_0$.
Shared ego observations every 0.5\,s, transitions, and planner random inputs then advance each state.

As shown in Fig.~\ref{fig:history_age}, historical roadside
evidence continues to benefit planning after communication
cutoff, with larger gains under ego-view occlusion.
A6 preserves positive planning gains longer than A4 and A5,
highlighting the benefits of persistent memory
and Source-Time Replay.

\begin{figure}[!t]
\centering
\captionsetup[subfigure]{font=small,skip=2pt,justification=centering,singlelinecheck=false}
\begin{subfigure}[t]{.49\columnwidth}
\centering
\includegraphics[width=\linewidth]{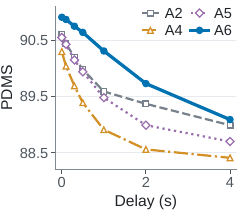}
\caption{Message delay}
\label{fig:async_delay}
\end{subfigure}\hfill
\begin{subfigure}[t]{.49\columnwidth}
\centering
\includegraphics[width=\linewidth]{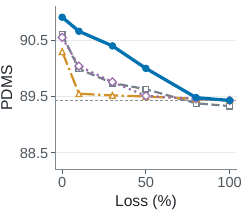}
\caption{Packet loss}
\label{fig:async_packet_loss}
\end{subfigure}
\caption{Planning under delayed delivery and packet loss.
A2: frame-wise cooperation; A4: arrival-time fusion; A5: age-aware fusion; A6: Source-Time Replay. (a) Additional delay relative to the original message timing; the 0\,s points match the fixed-weight ablation. (b) Packet loss at the original timing. The 2,390-anchor evaluation shares A6 parameters, ego inputs, initial states, local updates, and planner random inputs. The dotted reference is A1 (89.43 PDMS), reached by A4/A5/A6 under complete loss from stream start.}
\label{fig:asynchronous_robustness_curve}
\end{figure}

\paragraph{Communication delay and packet loss.}
Figure~\ref{fig:asynchronous_robustness_curve} evaluates
additional delays at fixed capture times and packet loss under the
original timing, sharing A6 weights and evaluation conditions across
A2/A4/A5/A6. Delays are $\{0,0.1,0.3,0.5,1,2,4\}$\,s and loss rates
are $\{0,10,30,50,80,100\}\%$. At rate $p$, packets drop independently with probability
$p/100$. Delay results average five shared inference seeds
with identical delay settings, initial states, and planner
random inputs across methods. Packet-loss results average
all $5\times5$ combinations of five network seeds (42--46)
and five inference seeds, sharing loss masks and planner
random inputs across methods for each combination.

Under complete message loss from stream start, A4/A5/A6 follow the shared ego update path and attain the A1 reference of 89.43 PDMS.
Cutoff tests retention, while stream-wide loss controls accumulated cooperative evidence.

Figure~\ref{fig:asynchronous_robustness_curve} shows
declining PDMS as delay and packet loss increase, while A6 maintains
higher scores under delayed delivery and partial loss.
These results highlight the robustness of Source-Time Replay
under asynchronous communication.

\section{Conclusion}
\label{sec:conclusion}
We presented \method, which unifies persistent cooperative memory,
planning-oriented communication, and world-conditioned trajectory
generation. Source-Time Replay propagates corrections
from delayed evidence to the current planning state.
Compact latent state tokens enable efficient roadside communication,
while predictive supervision promotes planning-relevant world
representations. \finalrev{Experiments demonstrate the planning benefits of
memory, replay, future supervision, and learned communication.}
\finalrev{Source-Time Replay reprocesses affected states after delayed
arrivals, potentially increasing computation under frequent arrivals.
Future work will explore incremental correction.}

\label{main-end}
\section*{LLM Usage Statement}
ChatGPT and Codex (OpenAI) assisted in refining author-written text,
related work, citations, and research ideas and claims, and generated
code, experiment-related content, figures, and data-related content.
The authors reviewed and approved these contributions and take full
responsibility for all submitted content.
\label{references-start}
\end{document}